\documentclass{article}
\usepackage[preprint]{neurips_2026}
\usepackage[utf8]{inputenc}
\usepackage[T1]{fontenc}
\usepackage{courier}
\usepackage{amsmath,amssymb}
\usepackage{booktabs}
\usepackage{graphicx}
\usepackage{url}
\usepackage{hyperref}
\setcitestyle{authoryear,square,semicolon}
\hypersetup{
  pdftitle={When Validation Stops Learning: Auditing Update Admission for Continual Embodied Agents},
  pdfauthor={Qinzhen Ma, Ruihai Wu},
  pdfsubject={Preprint},
  colorlinks=true,
  linkcolor=blue,
  citecolor=blue,
  urlcolor=blue
}

\title{When Validation Stops Learning: Auditing Update Admission for Continual Embodied Agents}
\author{
  Qinzhen Ma\\
  Rice University\\
  \texttt{qm18@rice.edu}
  \And
  Ruihai Wu\\
  University of California, Berkeley\\
  \texttt{ruihai@berkeley.edu}
}

\begin{document}
\maketitle
\begin{abstract}
Independent evaluation can reject harmful policy updates yet also prevent useful continual learning. We argue that update admission must be assessed through both error control and retained learning opportunities at a stated interaction budget. We identify a concrete failure: a range-based confidence gate cannot certify unchanged old-task behavior within otherwise substantial budgets. A standard paired-binomial construction reduces this burden when outcome disagreements are rare. We also specify certified historical-reference promotion and a round-level missed-opportunity metric. In a constructed one-step pushing diagnostic with 32 seeds, fresh paired checks admit 31.6\% of a common update stream at 2,000 episodes per stage, versus zero for the range-based gate; unconditional replay nevertheless learns better in closed-loop runs. A separate learned-dynamics stress test distinguishes model bias from feedback-selection error. The contribution is an admission-audit protocol with analytical and synthetic evidence; physical-robot and VLA validation remain open.
\end{abstract}

\section{Position: audit opportunities as well as accepted updates}
\label{sec:position}

A policy updated from demonstrations and corrections may forget earlier skills or exploit an inaccurate evaluator. Independent checks address one part of this problem. A gate that accepts nothing, however, can report no observed harmful admissions while its learner makes no progress. \textbf{Our position is that continual-learning evaluation must measure which useful update opportunities survive admission, alongside errors and the full cost of obtaining evidence.} We study policy updates; a world model screens candidates, and multiple agents are optional software organization.

\paragraph{Relation to prior work.}
Replay methods such as CLEAR mitigate forgetting during learning \citep{rolnick2019}; LIBERO makes sequential robot transfer measurable \citep{liu2023}. High Confidence Policy Improvement already supports confidence-based, incremental policy selection \citep{thomas2015}, and safe improvement under nonstationarity has also been studied \citep{chandak2020}. We do not introduce either idea. WorldEval assesses policies through generated outcomes \citep{li2025}; recent theory characterizes model exploitation \citep{bhamidipaty2026}. Our concern is the \emph{admission process after proposal and screening}, rather than a claim that proxy error is newly discovered. Adaptive holdout theory \citep{dwork2015} and evolving-agent systems \citep{zhang2025} further motivate separating development feedback from decision evidence.

Risk--coverage analysis is established in selective prediction \citep{geifman2017}; here the unit is a policy-update pool with historical obligations. Our contribution is a rule-specific feasibility analysis, an executable diagnostic exposing its learning consequences, and an opportunity audit that does not count normal candidate selection as an error. A tighter standard interval is a reference repair. The resulting negative findings matter: fewer observed violations need not mean a better continual learner.

\begin{figure}[t]
  \centering
  \includegraphics[width=\linewidth]{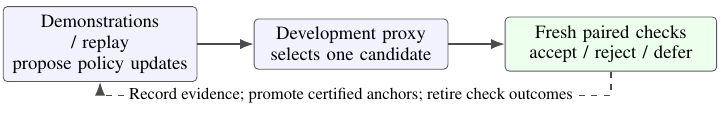}
  \caption{Evidence flow. Proposal and screening may reuse development feedback. Admission uses fresh environment pairs after commitment; all-candidate audit outcomes remain inaccessible to the learning loop.}
  \label{fig:flow}
\end{figure}

\section{An admission contract that can be checked within budget}
\label{sec:contract}

A task-conditioned policy $\pi_\theta$ retains its parameters across arriving task batches, with bounded demonstration replay and historical references. Thus adaptation can change earlier behavior; the process is continual policy learning rather than repeated prompting. For attempt $t$, commit a candidate $\pi'$, the incumbent $\pi$, task/reset distributions $P_j$, binary success $S$, and $m$ required comparisons. Set
\begin{equation}
g_j = \mathbb{E}_{P_j}[S(\pi')-S(b_j)],
\quad q_0 = \eta > 0,
\quad q_j = -\epsilon_j\ (j>0),
\quad g_j \geq q_j\ \forall j,
\label{eq:contract}
\end{equation}
where $b_0=\pi$ and historical $b_j$ are stored policies. Margins, coverage, and sample counts are fixed before checking. Model scores may rank a candidate pool but never count as independent environment outcomes. Different agents or fresh \emph{model-generated} seeds do not remove shared model bias.

\paragraph{A valid rule can freeze learning.}
For $n$ fresh pairs per comparison, $X=S(\pi')-S(b_j)\in\{-1,0,1\}$. Allocate $\alpha_t=6\delta/(\pi^2t^2)$. A simultaneous Hoeffding interval has radius $r_H=\sqrt{2\log(2m/\alpha_t)/n}$ \citep{hoeffding1963}. Even when every old-task pair has $X=0$, retention cannot pass unless $r_H\leq\epsilon$. With $m=3$, $t=1$, $\delta=\epsilon=0.05$, this requires 4,229 pairs per comparison. Equal allocation needs 25,374 episodes, exceeding a 20,000-episode cap. This is a limitation of this rule, not an information-theoretic lower bound.

\paragraph{Exploit paired disagreements, not just the outcome range.}
Let $p_+=\mathbb{P}(X=1)$ and $p_-=\mathbb{P}(X=-1)$, so $g=p_+-p_-$. Obtain ordinary Clopper--Pearson bounds $[\ell_+,u_+]$ and $[\ell_-,u_-]$ for their binomial counts, with each one-sided tail allocated $\beta=\alpha_t/(4m)$ \citep{clopper1934}. Use
\begin{equation}
L_j=\ell_+-u_-,\quad U_j=u_+-\ell_-,\quad
\text{accept if all } L_j\geq q_j;\quad
\text{reject if any } U_j<q_j;
\label{eq:admission}
\end{equation}
otherwise defer. Missing required evidence also means defer. When both counts are zero, $L=-(1-\beta^{1/n})$; retention needs only $n\geq\log\beta/\log(1-\epsilon)$, or 117 pairs in the example. This addresses the unchanged-skill failure; proving current-task gain can require more data. Allocate $n=\lfloor B/(2m)\rfloor$ under episode cap $B$, including both policies. Very small budgets still fail, and required evidence grows with task count and the attempt index.

Variance-aware alternatives already exist. We include an empirical Bernstein gate \citep{maurer2009} with the same error allocation and data, rather than treating all confidence gates as range-based. Paired binomial intervals apply to binary success; empirical Bernstein also permits general bounded scores.

\paragraph{Scope of assurance and historical references.}
Conditional on prior information, pairs must be i.i.d.\ within each committed task, with frozen policies and scoring. Union bounds over four tails, tasks, and attempts give probability at most $\delta$ of any accepted contract violation. Pair members may be correlated. We use fixed-size batches, no optional stopping, and fresh data/error allocation for any retry. This statement excludes future distribution drift, unprotected tasks, and physical hazards.

A permanent initial reference can miss later forgetting: a constructed history $0.30\to0.90\to0.26$ stays within $0.05$ of the initial policy. Instead, after acceptance, replace a historical reference only if its same valid interval has $L_j\geq0$. On the coverage event, reference performance never decreases on its fixed distribution. This protects \emph{certified improvements}, not an unknown historical optimum. Cap the active reference set; eviction ends protection. Log policy/data hashes and decisions; anomalies can trigger checkpoint restoration without implying safety under drift.

\begin{figure}[t]
  \centering
  \includegraphics[width=\linewidth]{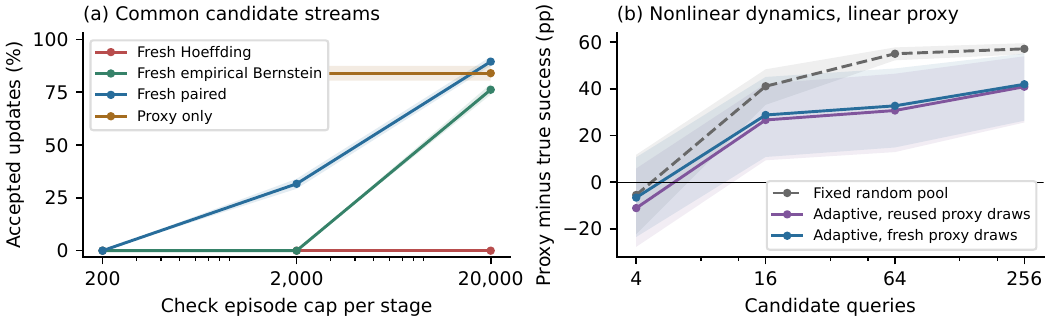}
  \caption{Executed diagnostic simulations. (a) Identical candidate streams and references isolate the checking rule. (b) More proxy optimization can expose dynamics error; fresh proxy draws do not eliminate it. Bands are 95\% percentile bootstrap intervals over 32 seeds (4,000 resamples). Further details are provided in Appendices C and D.}
  \label{fig:results}
\end{figure}

\begin{figure}[ht]
  \hrule
  \vspace{3pt}
  \noindent\textbf{Algorithm 1. One committed update attempt}\par
  \vspace{3pt}
  \begin{tabular}{@{}r@{\quad}p{0.92\linewidth}@{}}
  1 & Generate candidates; select one using development evidence.\\
  2 & Commit candidate, references, required tasks, counts, and $\alpha_t$.\\
  3 & If a required check is unfunded: \textsc{Defer}; otherwise draw fresh pairs.\\
  4 & All $L_j\geq q_j$: \textsc{Accept}; any $U_j<q_j$: \textsc{Reject}; else \textsc{Defer}.\\
  5 & On acceptance: adopt; initialize new references; promote old ones if $L_j\geq0$.\\
  6 & Log evidence; retire check outcomes; advance the attempt index.
  \end{tabular}
  \vspace{3pt}
  \hrule
\end{figure}

\section{Executed diagnostics and an opportunity-level audit}
\label{sec:diagnostics}

\paragraph{A metric with the correct unit of choice.}
Let $E_t$ indicate that at least one candidate in the committed pool satisfies every true contract threshold. Define missed opportunity as an $E_t$ round in which no satisfying candidate is adopted, divided by the number of $E_t$ rounds. A pool of four valid candidates and one adoption therefore has zero misses, whereas a candidate-level ``valid but not adopted'' rate gives a misleading 75\%. Report coverage, contract violations per admission, retention violations separately, new-task gain, forgetting, and all resource costs; zero denominators are undefined.

\paragraph{Sequential pushing diagnostic.}
We executed a deliberately simple, one-step simulator: displacement $g_j(1+z)a$, action $a=\operatorname{clip}[(\theta_{\mathrm{shared}}+\theta_j)d,0,3]$, desired displacement $d$, and success error at most $0.06$. Ten mobility tasks arrive sequentially; two initialize the policy, followed by eight updates. Synthetic demonstrations produce four candidates: two current-task ridge fits, a replay fit, and a task-local fit. A linear dynamics proxy is fitted from separate simulated transitions. The local candidate preserves old behavior, intentionally testing the low-disagreement regime. This is not a contact-rich robot or vision benchmark.

Each task has a separate 4,096-reset audit population. Checks sample fresh indices with replacement after commitment; after the stream, exhaustive evaluation labels every candidate exactly \emph{on this finite population}. No audit labels train or select policies. Exhaustion costs more than the largest checking cap and is charged separately. Thirty-two fixed evaluation seeds, disjoint from four pilot seeds, vary task order, data, and resets. Common candidate streams isolate gates; closed-loop runs let decisions change later proposals. All gates receive four candidates and matched episode caps; unguarded baselines leave unused checking budget unspent. Consequently, these are \emph{admission diagnostics}, not claims of resource-optimal policy learning.

\paragraph{What the results establish---and do not.}
On common streams (Fig.~\ref{fig:results}a), the paired gate admits 81/256 updates at $B=2{,}000$ and 229/256 at $B=20{,}000$, with no observed contract violations; Hoeffding admits none. Empirical Bernstein admits 0 and 195/256, respectively, also without observed violations. All three defer throughout at $B=200$. Thus the standard interval choice changes whether valid update opportunities survive. This does not establish zero population risk or a universally efficient rule.

\begin{table}[t]
  \caption{Closed-loop diagnostic, $B=2{,}000$, 32 seeds. Values are percentages; forgetting is percentage points. Violations include insufficient gain, not just harmful regression. ``--'' means no admissions. Uncertainty and counts are in Appendix~D.}
  \label{tab:closedloop}
  \centering
  \begin{tabular}{lrrrrr}
    \toprule
    Rule & Adopt & Violate & Miss & Final success & Forget\\
    \midrule
    Frozen policy & 0.0 & -- & 100.0 & 31.4 & 0.0\\
    Direct fine-tuning & 100.0 & 99.6 & 99.4 & 28.1 & 79.9\\
    Replay update & 100.0 & 5.9 & 0.0 & 100.0 & 0.0\\
    Proxy only & 87.1 & 2.7 & 10.0 & 98.4 & 0.5\\
    Fresh Hoeffding & 0.0 & -- & 100.0 & 31.4 & 0.0\\
    Fresh emp.\ Bernstein & 0.0 & -- & 100.0 & 31.4 & 0.0\\
    Fresh paired & 37.1 & 0.0 & 60.4 & 59.6 & 0.0\\
    Paired, no proxy & 37.5 & 0.0 & 58.4 & 59.8 & 0.0\\
    \bottomrule
  \end{tabular}
\end{table}

Closed-loop results expose the remaining opportunity cost (Table~\ref{tab:closedloop}): at $B=2{,}000$, paired checks miss 60.4\% of available opportunities, and replay obtains higher final success. Replay's 5.9\% contract violations are insufficient current-task gains, not retention losses. Removing the proxy slightly improves paired-gate coverage; these data do not justify multi-agent orchestration or proxy screening as necessary components.

\paragraph{Separate model bias from selection feedback.}
A second diagnostic fits a linear proxy to low-action transitions from $x'=a-ca^2$, then compares fixed random and adaptive candidate search at equal query counts, using reused or fresh proxy draws. Proposal and feedback random streams are separate. At $c=0.35$ and 256 adaptive queries, held-out proxy success exceeds true success by 40.8 percentage points with reused draws and 41.9 with fresh draws; at $c=0$, the corresponding gaps are 0.15 and 0.32 points. Selection-score optimism persists even with fresh draws. This distinguishes \emph{fresh proxy feedback} from independent evidence about the true environment; it does not establish that adaptive search always exploits models more than random search.

\section{Implications and limits}
\label{sec:limits}

Reliable continual learning requires a gate with useful power, references that reflect certified acquired skills, and an audit that counts opportunities correctly. The evidence here supports these diagnostic requirements, not superiority over replay. The simulator has known task identity, easy local adapters, and exact finite-population audits; shared VLA updates may have far more outcome disagreement. A structural proof that an update cannot change old behavior could bypass those checks; our diagnostic treats policies as black-box evaluands. Reliable physical resets, distribution drift, and affordable broad task coverage remain unresolved. A next test should transplant the same opportunity audit and matched candidate streams to LIBERO and then physical tasks. Failure to improve the error--opportunity trade-off under those conditions would limit the practical case for this admission design.

\clearpage
\appendix
\section{Statistical construction and reference promotion}
\label{app:statistical}

Condition on all information available before attempt $t$, including the adaptively generated candidate, protected-task set, reference policies, scoring, sample counts, and reset distributions. For comparison $j$, let the $n_j$ paired binary differences be i.i.d., and let $k_+$ and $k_-$ count positive and negative differences. Each count is marginally binomial, even though the two counts are dependent. For a count $k$ in $n$ trials and tail probability $\beta$, define
\[
\ell(k)=
\begin{cases}
0 & k=0,\\
F^{-1}_{\operatorname{Beta}(k,n-k+1)}(\beta) & k>0,
\end{cases}
\qquad
u(k)=
\begin{cases}
1 & k=n,\\
F^{-1}_{\operatorname{Beta}(k+1,n-k)}(1-\beta) & k<n.
\end{cases}
\]
Binomial-tail inversion gives each one-sided noncoverage probability at most $\beta$. With $\beta=\alpha_t/(4m_t)$, a union bound over the four endpoints and $m_t$ comparisons gives simultaneous coverage at least $1-\alpha_t$. On that event,
\[
\ell(k_+)-u(k_-)\le p_+-p_-\le u(k_+)-\ell(k_-).
\]
No independence between endpoints or tasks is required. Taking expectations over prior information, followed by a countable union bound over attempts, yields failure probability at most $\sum_{t\ge1}6\delta/(\pi^2t^2)=\delta$. Acceptance on the coverage event implies every committed contract holds. This is a composition of standard intervals, not a new concentration result. The software additionally enumerates exact multinomial outcome probabilities at several parameter settings to check coverage numerically; that check is not a substitute for the argument.

\paragraph{Reference promotion.}
Initialize a task's reference upon admission (and initialize warm-up references before the stream). After accepting a candidate, promote an old reference only when its existing lower bound is nonnegative. On the same simultaneous coverage event, the promoted policy has expected success at least that of the previous reference. Induction makes the reference sequence nondecreasing for a fixed $P_j$. Future accepted policies lie within the declared $\epsilon_j$ of the current reference. The guarantee concerns this certified sequence, not the best-ever policy: a real improvement too small to certify need not be incorporated. Promotion consumes no second batch because the existing simultaneous event already covers the needed comparison. Changing a task distribution changes the meaning of the claim and requires a new logged contract.

The code promotes references in closed-loop guarded runs. The constructed tasks often reach near-perfect success upon admission, so these experiments do not measure the practical benefit of reference promotion. The $0.30\to0.90\to0.26$ example is a stipulated counterexample to permanent initial references, not measured robot performance.

\paragraph{Rule-specific feasibility.}
For zero observed disagreements, both binomial upper endpoints equal $1-\beta^{1/n}$ and both lower endpoints are zero. Therefore $L\ge-\epsilon$ exactly when $n\ge\log\beta/\log(1-\epsilon)$. The Hoeffding comparison instead needs $n\ge2\log(2m/\alpha_t)/\epsilon^2$. For $(m,t)=(3,1)$ the minimum integer counts are 117 and 4,229; for $(10,8)$ they are 222 and 8,519. With equal allocation, the corresponding episode costs for the latter are 4,440 and 170,380. These counts concern an old-task zero-disagreement check. They neither guarantee acceptance of a current-task gain nor imply comparable efficiency when policies frequently disagree. Both rules can eventually outgrow a fixed budget as protected tasks or verification attempts increase.

\paragraph{Empirical Bernstein control.}
Rescale Theorem 4 of \citet{maurer2009} from $[0,1]$ to $[-1,1]$ and assign each tail probability $\alpha_t/(2m)$. For unbiased sample variance $s_j^2$, the resulting radius is
\[
r_{\mathrm{EB},j}=
\sqrt{\frac{2s_j^2\log(4m/\alpha_t)}{n}}
+\frac{14\log(4m/\alpha_t)}{3(n-1)},\qquad n\ge2.
\]
Apply the same three-way rule to the clipped interval around the pair mean; defer at $n<2$. This baseline also becomes more informative for rare disagreements. Its constants and its admission coverage differ from exact binomial inversion; neither construction is claimed novel or uniformly best.

\paragraph{Operational conditions.}
The supplied rule supports binary scores only. Physical state carry-over, correlated reset pairs, uncommitted sample stopping, or changing a policy/scoring function during a batch invalidate the stated assumptions. Reusing a prior check in a later batch is not allowed. A confidence-sequence design could permit sequential sampling, but is not implemented. Restoring a checkpoint is a recovery action, not a guarantee that either policy remains suitable after environmental drift. The active reference capacity is ten tasks in this experiment; a deployment with a larger stream must declare eviction and its lost coverage. Decision logs grow with attempts and must also be counted as storage.

\section{Executable diagnostic specification}
\label{app:diagnostic}

All demonstrations, dynamics observations, policy updates, and evaluations in this submission are synthetic. There are no human participants, proprietary data, pretrained VLAs, LIBERO runs, learned video models, or physical-robot trials. The original diagnostic runner is \texttt{experiment.py}; \texttt{admission.py} contains the interval logic, and \texttt{summarize.py} creates the figures and table from recorded outputs. The experimental implementation, configuration files, source hashes, commands, raw decisions, and pilot runs are not included in this preprint's LaTeX source package.

\paragraph{Environment and data.}
For each run, permute ten nominal mobilities evenly spaced from 0.55 to 1.6. A reset has desired displacement $d\sim\operatorname{Uniform}(0.5,1)$ and mobility perturbation $z\sim\operatorname{Uniform}(-0.04,0.04)$. The single action is $a=\operatorname{clip}[(\theta_{\mathrm{shared}}+\theta_j)d,0,3]$. Terminal displacement is $g_j(1+z)a$; success is absolute target error at most 0.06. The task descriptor is its known integer identity. This quasi-static scalar model omits contact geometry, inertia, perception, and multi-step planning.

For each task generate 32 analytic-demonstrator actions $a^*=d/[g_j(1+z)]+e$, with $e\sim\mathcal{N}(0,0.01^2)$. Their normalized least-squares coefficient is $w_j^*=\sum_i d_i a_i^*/\sum_i d_i^2$. Two tasks initialize $\theta_{\mathrm{shared}}$ as the average of their fits, with their task offsets adjusted to reproduce each fit; other offsets start at zero. Thereafter parameters persist through eight arriving tasks. Replay stores the released synthetic demonstrations, with capacity 320 episodes, sufficient for this ten-task stream.

Independently fit a per-task linear proxy $\widehat{x}'=\widehat{g}_j a$ from 32 probes with $a\sim\operatorname{Uniform}(0.05,0.2)$ and observed displacement $g_j a+e$, $e\sim\mathcal{N}(0,0.03^2)$. These 320 pretraining probes include all task identities and are shared across compared rules. The proxy ignores reset perturbation $z$ and remains frozen. This pretraining access is a simplifying assumption, not evidence of unseen-task world-model generalization. A distinct development bank has 64 resets per task. The audit bank has 4,096 independently generated resets per task and defines the finite distribution to which the guarantee applies.

\paragraph{Candidate generation and replay.}
Let $X_j$ be the feature row with a shared coordinate and the task's offset coordinate equal to one. Two candidates minimize normalized current-task imitation loss plus $\lambda\lVert\theta-\theta_{\mathrm{parent}}\rVert^2$, for $\lambda\in\{0.01,0.1\}$. A third minimizes a 50/50 mixture of current-task and uniform historical imitation losses with $\lambda=0.01$. Each is computed by a strictly positive-definite ridge solve. The fourth is an exact task-local least-squares update holding the shared coordinate fixed; it leaves previous task actions unchanged. All four candidates are supplied to each compared rule. The easy local candidate deliberately provides a low-disagreement control; this makes the diagnostic unsuitable for claims about general policy capacity or state-of-the-art robot learning.

\paragraph{Selection and admission.}
Proxy methods select the candidate maximizing $\min_j(\widehat{g}_j-q_j)$ on development resets, breaking ties by candidate index. The proxy-only baseline adopts when this estimated minimum slack is nonnegative. The three guarded proxy methods use the same proxy-selected candidate and differ only in their interval. The no-proxy variant selects by the same 50/50 development imitation/replay loss. Direct fine-tuning adopts the first current-task candidate; replay adopts the replay-trained candidate; the frozen baseline ignores updates. These unconditional baselines may adopt a no-op on an already-solved current task, violating the positive-gain contract without causing forgetting.

At each attempt use $\eta=\epsilon=\delta=0.05$, caps $B\in\{200,2000,20000\}$, and $n=\lfloor B/(2m)\rfloor$. The verifier draws independent uniform audit-bank indices with replacement only after candidate selection. Count both candidate and reference episodes; no retries or optional stopping occur. Historical protection covers admitted tasks and starts with the two warm-up tasks. In common-stream diagnostics, a source learner always advances with the local candidate and retains the source references; every gate is judged against this same externally fixed contract stream. Its decisions do not alter the source. Closed-loop runs instead branch each learner and update its own references after acceptance. The three guarded proxy methods differ only in their interval construction.

\paragraph{Audit and uncertainty.}
After a stream ends, exhaustively evaluate every proposed candidate against its associated incumbent and historical references. Thus all contract labels are exact for the committed finite bank, including screen-outs and deferrals; no audit confidence interval or unresolved category is needed here. On a non-enumerable environment, a separate sufficiently powered audit and explicit unresolved labels would still be required. The audit costs $2KmN$ outcome evaluations per stage for $K=4$ candidates and $N=4096$ resets, with repeated references counted conservatively. The main learning loop receives none of these exhaustive results.

Coverage is admissions divided by eight opportunities to update. A contract violation means any $g_j<q_j$; retention violations are reported separately. A missed opportunity requires at least one contract-satisfying candidate and no satisfying adoption. Final success averages all ten tasks. Forgetting averages, over the first nine tasks, the maximum success after that task's arrival minus final success, using exact bank evaluations of all deployed checkpoints. We report percentile bootstrap uncertainty by resampling the 32 entire seed trajectories, not individual dependent candidates, 4,000 times with bootstrap seed 20260908. Zero observed violations are not asserted to establish a zero error rate.

\paragraph{Development and evaluation separation.}
Pilot seeds 0--3 were used to check the implementation. Before evaluating seeds 100--131, we increased the audit bank from 128 to 4,096 to prevent cheap exhaustive checking inside the stated caps, replaced heuristic update fractions with explicit ridge fits, separated harmful regressions from minimum-gain violations, and corrected proxy-cost accounting. Subsequent implementation review found coupling between proposal and feedback random streams in the proxy test; we separated them and reran the full fixed seed set. We also added the standard empirical Bernstein control as a robustness comparison. The original study retained earlier outputs and dated change records; these are not included in this preprint package. No candidate, noise, threshold, or budget hyperparameters were tuned on the evaluation outcomes. The bootstrap and plots do not feed back into a learner.

\section{Factorial proxy-feedback stress test}
\label{app:proxy-stress}

This is a separate one-step environment, not a closed-loop continual-learning benchmark. True displacement is $x'=a-ca^2$, with $c\in\{0,0.35\}$. Fit a linear model through the origin using 32 actions uniform on $[0.05,0.3]$, with Gaussian observation noise of standard deviation 0.01. Search constant-action policies in $[0.2,1.6]$ for targets $0.65+u$, where $u\sim\operatorname{Uniform}(-0.08,0.08)$ and success tolerance is 0.06.

At query budgets $K\in\{4,16,64,256\}$, random search predefines independent uniform actions; adaptive search begins with the same random first action and subsequently perturbs the best-so-far action with Gaussian standard deviation 0.15, clipping to the action range. Both choose the largest observed proxy success, with first occurrence resolving ties. One feedback condition reuses 16 development target offsets; the other draws 16 new offsets per query. A final 128-point target grid, inaccessible to selection, provides exact paired proxy/true scores for the chosen action. The full $2\times2\times2\times4$ design runs on each of 32 seeds, producing 1,024 recorded searches.

We distinguish model gap (held-out proxy score minus true score) from selection gap (winning development score minus held-out proxy score). Fresh per-query feedback does not make the winning score an independent estimate: selection still favors favorable estimates. With curvature 0.35, the adaptive 256-query model gap is 40.8 points [25.5,53.9] for reused feedback and 41.9 [26.3,55.1] for fresh feedback. With curvature zero, the gaps are 0.15 [-0.42,0.76] and 0.32 [-0.05,0.78]. These are bootstrap intervals across seeds, not physical-robot uncertainty. Selection gaps under curvature 0.35 are 11.5 [8.2,14.8] and 19.3 [15.8,22.4] points, respectively. The results do not support the claim that simply refreshing proxy draws removes selection optimism or dynamics bias.

\section{Counts, uncertainty, and resource reporting}
\label{app:counts}

\paragraph{Closed-loop uncertainty.}
Values below are seed-cluster bootstrap intervals at $B=2{,}000$. Violation counts are exact on the finite audit populations; zeros are observations, not a claim of zero risk.

\begin{center}
\small
\begin{tabular}{@{}lrrr@{}}
\toprule
Rule & Coverage (\%) & Final success (\%) & Forgetting (pp) \\
\midrule
Frozen policy & 0.0 [0.0, 0.0] & 31.4 [29.5, 33.2] & 0.0 [0.0, 0.0] \\
Direct fine-tuning & 100.0 [100.0, 100.0] & 28.1 [24.4, 31.8] & 79.9 [75.8, 84.0] \\
Replay update & 100.0 [100.0, 100.0] & 100.0 [100.0, 100.0] & 0.0 [0.0, 0.0] \\
Proxy only & 87.1 [83.6, 90.6] & 98.4 [96.8, 99.5] & 0.5 [0.0, 1.6] \\
Fresh Hoeffding & 0.0 [0.0, 0.0] & 31.4 [29.5, 33.2] & 0.0 [0.0, 0.0] \\
Fresh emp.\ Bernstein & 0.0 [0.0, 0.0] & 31.4 [29.5, 33.2] & 0.0 [0.0, 0.0] \\
Fresh paired & 37.1 [36.3, 37.5] & 59.6 [57.4, 61.8] & 0.0 [0.0, 0.0] \\
Paired, no proxy & 37.5 [37.5, 37.5] & 59.8 [58.0, 61.6] & 0.0 [0.0, 0.0] \\
\bottomrule
\end{tabular}
\end{center}

\begin{center}
\small
\begin{tabular}{@{}lrrr@{}}
\toprule
Rule & Violate / adopt & Old violations & Miss / available \\
\midrule
Frozen policy & 0/0 & 0 & 233/233 \\
Direct fine-tuning & 255/256 & 251 & 154/155 \\
Replay update & 15/256 & 0 & 0/241 \\
Proxy only & 6/223 & 6 & 24/241 \\
Fresh Hoeffding & 0/0 & 0 & 233/233 \\
Fresh emp.\ Bernstein & 0/0 & 0 & 233/233 \\
Fresh paired & 0/95 & 0 & 145/240 \\
Paired, no proxy & 0/96 & 0 & 135/231 \\
\bottomrule
\end{tabular}
\end{center}

\paragraph{Common-stream counts.}
Each configuration has 256 attempted update rounds and 233 rounds with at least one feasible candidate. Hoeffding accepts zero at every tested cap. Paired checking accepts 0, 81, and 229 at caps 200, 2,000, and 20,000, respectively. The corresponding missed-opportunity counts are 233, 152, and 4. Empirical Bernstein accepts 0, 0, and 195, with zero observed violations. Proxy-only checking accepts 215 with one contract violation, zero retention violations, and 19 missed opportunities; its decisions do not use the check cap. The no-proxy paired variant accepts 0, 84, and 232 with zero observed violations.

\paragraph{Resource accounting.}
Every candidate is an explicit small ridge solve or local fit; no GPU is used. The complete recorded evaluation contains 1,536 seed/budget/method/view settings, 12,288 decisions, and 1,024 proxy-search settings, taking 50.9 seconds in the reported CPU environment including exhaustive audits. This runtime describes scalar diagnostic simulation and cannot estimate robot or VLA cost. Teacher data and 320 dynamics pretraining probes per seed are shared across methods. The unguarded baselines use no admission episodes; they are not artificially forced to spend their allowance. No claim of practical computational advantage is made.

\begin{center}
\small
\begin{tabular}{@{}lrrr@{}}
\toprule
Rule & Check episodes & Proxy predictions & Candidate audit \\
\midrule
Frozen policy & 0 & 0 & 786,432 \\
Direct fine-tuning & 0 & 0 & 1,703,936 \\
Replay update & 0 & 0 & 1,703,936 \\
Proxy only & 0 & 24,368 & 1,559,552 \\
Fresh Hoeffding & 15,984 & 12,288 & 786,432 \\
Fresh emp.\ Bernstein & 15,984 & 12,288 & 786,432 \\
Fresh paired & 15,963 & 20,832 & 1,333,248 \\
Paired, no proxy & 15,961 & 0 & 1,348,608 \\
\bottomrule
\end{tabular}
\end{center}

Counts are means per stream, rounded to whole outcomes. Candidate audits exclude $9\times10\times4096=368{,}640$ additional checkpoint outcomes per stream for final success and forgetting. The reported raw outputs cover all budgets, decisions, candidate/reference gains, and the full proxy-feedback factorial; those files are not included in this preprint package. Replay never replaces fresh checks.


\clearpage
\begin{thebibliography}{12}

\bibitem[Bhamidipaty et~al.(2026)]{bhamidipaty2026}
Logan Mondal Bhamidipaty, Esmeralda S. Whitammer, David Abel, Mykel J. Kochenderfer, and Subramanian Ramamoorthy.
\newblock Imperfect world models are exploitable.
\newblock \emph{arXiv preprint arXiv:2605.15960}, 2026.
\newblock URL \url{https://arxiv.org/abs/2605.15960}.

\bibitem[Chandak et~al.(2020)]{chandak2020}
Yash Chandak, Scott M. Jordan, Georgios Theocharous, Martha White, and Philip S. Thomas.
\newblock Towards safe policy improvement for non-stationary MDPs.
\newblock In \emph{Advances in Neural Information Processing Systems}, volume 33, 2020.
\newblock URL \url{https://arxiv.org/abs/2010.12645}.

\bibitem[Clopper and Pearson(1934)]{clopper1934}
C. J. Clopper and E. S. Pearson.
\newblock The use of confidence or fiducial limits illustrated in the case of the binomial.
\newblock \emph{Biometrika}, 26(4):404--413, 1934.
\newblock doi: 10.1093/biomet/26.4.404.
\newblock URL \url{https://academic.oup.com/biomet/article/26/4/404/291538}.

\bibitem[Dwork et~al.(2015)]{dwork2015}
Cynthia Dwork, Vitaly Feldman, Moritz Hardt, Toniann Pitassi, Omer Reingold, and Aaron Roth.
\newblock The reusable holdout: Preserving validity in adaptive data analysis.
\newblock \emph{Science}, 349(6248):636--638, 2015.
\newblock doi: 10.1126/science.aaa9375.
\newblock URL \url{https://www.cis.upenn.edu/~aaroth/reusable.html}.

\bibitem[Geifman and El-Yaniv(2017)]{geifman2017}
Yonatan Geifman and Ran El-Yaniv.
\newblock Selective classification for deep neural networks.
\newblock In \emph{Advances in Neural Information Processing Systems}, volume 30, 2017.
\newblock URL \url{https://papers.nips.cc/paper/7073-selective-classification-for-deep-neural-networks}.

\bibitem[Hoeffding(1963)]{hoeffding1963}
Wassily Hoeffding.
\newblock Probability inequalities for sums of bounded random variables.
\newblock \emph{Journal of the American Statistical Association}, 58(301):13--30, 1963.
\newblock doi: 10.1080/01621459.1963.10500830.
\newblock URL \url{https://www.tandfonline.com/doi/abs/10.1080/01621459.1963.10500830}.

\bibitem[Li et~al.(2025)]{li2025}
Yaxuan Li, Yichen Zhu, Junjie Wen, Chaomin Shen, and Yi Xu.
\newblock WorldEval: World model as real-world robot policies evaluator.
\newblock \emph{arXiv preprint arXiv:2505.19017}, 2025.
\newblock URL \url{https://arxiv.org/abs/2505.19017}.

\bibitem[Liu et~al.(2023)]{liu2023}
Bo Liu, Yifeng Zhu, Chongkai Gao, Yihao Feng, Qiang Liu, Yuke Zhu, and Peter Stone.
\newblock LIBERO: Benchmarking knowledge transfer for lifelong robot learning.
\newblock In \emph{Advances in Neural Information Processing Systems}, volume 36, pages 44776--44791, 2023.
\newblock URL \url{https://arxiv.org/abs/2306.03310}.

\bibitem[Maurer and Pontil(2009)]{maurer2009}
Andreas Maurer and Massimiliano Pontil.
\newblock Empirical bernstein bounds and sample variance penalization.
\newblock \emph{arXiv preprint arXiv:0907.3740}, 2009.
\newblock URL \url{https://arxiv.org/abs/0907.3740}.

\bibitem[Rolnick et~al.(2019)]{rolnick2019}
David Rolnick, Arun Ahuja, Jonathan Schwarz, Timothy P. Lillicrap, and Greg Wayne.
\newblock Experience replay for continual learning.
\newblock In \emph{Advances in Neural Information Processing Systems}, volume 32, 2019.
\newblock URL \url{https://papers.nips.cc/paper_files/paper/2019/hash/fa7cdfad1a5aaf8370ebeda47a1ff1c3-Abstract.html}.

\bibitem[Thomas et~al.(2015)]{thomas2015}
Philip Thomas, Georgios Theocharous, and Mohammad Ghavamzadeh.
\newblock High confidence policy improvement.
\newblock In \emph{Proceedings of the 32nd International Conference on Machine Learning}, volume 37 of \emph{Proceedings of Machine Learning Research}, pages 2380--2388. PMLR, 2015.
\newblock URL \url{https://proceedings.mlr.press/v37/thomas15.html}.

\bibitem[Zhang et~al.(2025)]{zhang2025}
Jenny Zhang, Shengran Hu, Cong Lu, Robert Lange, and Jeff Clune.
\newblock Darwin G\"odel machine: Open-ended evolution of self-improving agents.
\newblock \emph{arXiv preprint arXiv:2505.22954}, 2025.
\newblock URL \url{https://arxiv.org/abs/2505.22954}.

\end{thebibliography}
\end{document}